\documentclass[letterpaper]{article} % DO NOT CHANGE THIS
\usepackage{aaai25}  % DO NOT CHANGE THIS
\usepackage{times}  % DO NOT CHANGE THIS
\usepackage{helvet}  % DO NOT CHANGE THIS
\usepackage{courier}  % DO NOT CHANGE THIS
\usepackage[hyphens]{url}  % DO NOT CHANGE THIS
\usepackage{graphicx} % DO NOT CHANGE THIS
\usepackage{natbib}  % DO NOT CHANGE THIS AND DO NOT ADD ANY OPTIONS TO IT
\usepackage{caption} % DO NOT CHANGE THIS AND DO NOT ADD ANY OPTIONS TO IT
\usepackage{algorithm}
\usepackage{algorithmic}

\usepackage{newfloat}
\usepackage{listings}
\DeclareCaptionStyle{ruled}{labelfont=normalfont,labelsep=colon,strut=off} % DO NOT CHANGE THIS
\floatstyle{ruled}
\newfloat{listing}{tb}{lst}{}
\floatname{listing}{Listing}
\usepackage{booktabs}
\usepackage{amsmath}
\usepackage{xcolor}
\usepackage{amssymb}

\title{Universal Domain Adaptive Object Detection via Dual Probabilistic Alignment}
\author{
    Yuanfan Zheng\textsuperscript{\rm 1,2}\equalcontrib,
    Jinlin Wu\textsuperscript{\rm 1,2}\equalcontrib,
    Wuyang Li\textsuperscript{\rm 3},
    Zhen Chen\textsuperscript{\rm 1}\thanks{Corresponding author.}
}
\affiliations{
    \textsuperscript{\rm 1}CAIR, HKISI-CAS  \,\,\,\, 
    \textsuperscript{\rm 2}MAIS, Institute of Automation, Chinese Academy of Sciences\\
    \textsuperscript{\rm 3}The Chinese University of Hong Kong\\
    zyfone.cs@gmail.com, jinlin.wu@nlpr.ia.ac.cn, wymanbest@outlook.com, zchen.francis@gmail.com
}

\usepackage{bibentry}
\begin{document}

\maketitle

\begin{abstract}
Domain Adaptive Object Detection (DAOD) transfers knowledge from a labeled source domain to an unannotated target domain under closed-set assumption. Universal DAOD (UniDAOD) extends DAOD to handle open-set, partial-set, and closed-set domain adaptation. In this paper, we first unveil two issues: domain-private category alignment is crucial for global-level features, and the domain probability heterogeneity of features across different levels. To address these issues, we propose a novel Dual Probabilistic Alignment (DPA) framework to model domain probability as Gaussian distribution, enabling the heterogeneity domain distribution sampling and measurement. The DPA consists of three tailored modules: the Global-level Domain Private Alignment (GDPA), the Instance-level Domain Shared Alignment (IDSA), and the Private Class Constraint (PCC). GDPA utilizes the global-level sampling to mine domain-private category samples and calculate alignment weight through a cumulative distribution function to address the global-level private category alignment. IDSA utilizes instance-level sampling to mine domain-shared category samples and calculates alignment weight through Gaussian distribution to conduct the domain-shared category domain alignment to address the feature heterogeneity. The PCC aggregates domain-private category centroids between feature and probability spaces to mitigate negative transfer. Extensive experiments demonstrate that our DPA outperforms state-of-the-art UniDAOD and DAOD methods across various datasets and scenarios, including open, partial, and closed sets.  Codes are available
 at \url{https://github.com/zyfone/DPA}.
\end{abstract}


%     \link{Code}{https://github.com/zyfone/DPA}

\section{Introduction}
Object detection has made significant progress in recent years \cite{li2021htd,jia2023detrs,zhao2024detrs}. However, the well-trained object detector failed to generalize in novel domain scenarios due to domain shift. Domain Adaptive Object Detection (DAOD) \cite{krishna2023mila,huang2024blenda} transferring from the source domain to the unlabelled target domain to overcome domain shift and has been widely applied in medical analysis \cite{pu2024m3, ali2024assessing,liu2023decoupled}, autonomous driving \cite{cai2024uncertainty,shi2024domain} and robotic understanding \cite{chapman2023predicting,Li2024cliff}. However, DAOD is limited by the closed-set assumption \cite{ben2010theory,li2023adjustment} failing to generalize to real-world scenarios. To address this, Universal DAOD (UniDAOD) endows DAOD with open-set domain adaption capabilities to overcome label shifts without prior knowledge of categories. The former work, US-DAF \cite{shi2022universal}, leverages the threshold filter mechanism and the scale-sensitive domain alignment. CODE \cite{shi2024confused} adopts virtual domain alignment to avoid aligning domain-private category samples to mitigate negative transfer. Other methods \cite{lang2023class,shi2024dynamically} adopt dynamic weighting for domain-private categories to facilitate positive transfer. Essentially, existing UniDAOD approaches align shared categories at both the global and instance levels while ignoring the alignment of domain-private categories.

\begin{figure}[tb]
\centering
\includegraphics[width=0.45\textwidth]{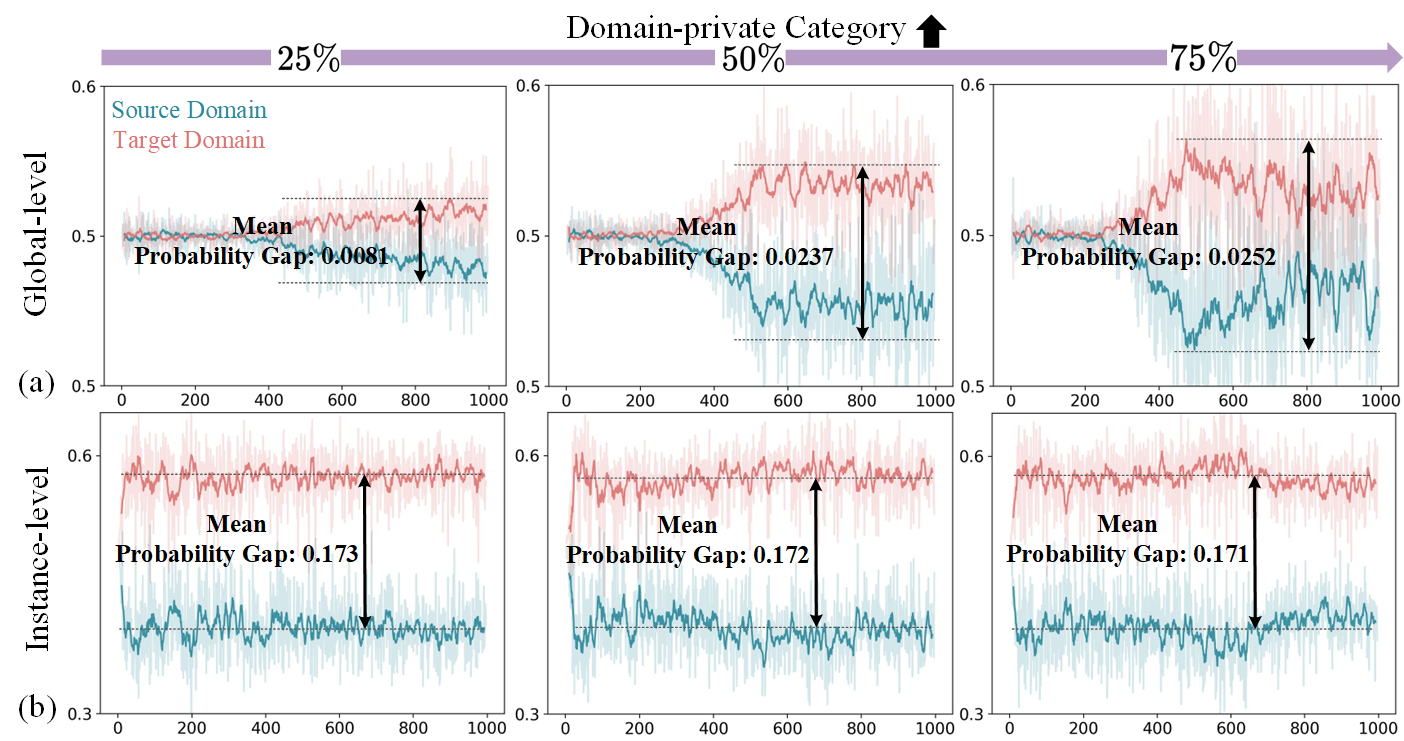}
\caption{
The visualization of domain probability in the domain discriminator\footnote{}. (a) As the number of domain-private categories increases, a more distinct gap emerges between the two domains at the global level, suggesting the \textbf{alignment of the domain-private category}. (b) With the increases of the domain-private category, the mean probability gap remains approximately constant at the instance level, indicating the \textbf{alignment of domain-shared category}.} 
\label{domain-probs}
\end{figure}
\footnotetext{The horizontal axis is training iteration ($\times$100), and the vertical axis is the probability of domain discriminator.}
\par
Despite significant progress in recent years, the current UniDAOD paradigm encounters two major issues that result in suboptimal domain alignment. The first issue is that they overlook global feature alignment with domain-private categories. Due to the agnostic prior category knowledge, existing methods \cite{lang2023class,shi2024dynamically} primarily focus on estimating the domain-shared category set to mitigate negative transfer, \textbf{wrongly assuming} that both global and instance features fairly contribute to the domain-shared alignment. For the first time, we empirically reveal the issue of this assumption in Fig. \ref{domain-probs}. We claim \emph{the fact is that global-level features tend to align domain private categories, while instance-level features tend to align domain shared categories}, which can also be thoroughly justified in Fig. \ref{last_prob}. This phenomenon motivates us to revisit the UniDAOD domain alignment, focusing on the domain-private category alignment at the global level for UniDAOD.
\par
The second issue is the heterogeneity of features at different feature levels. Since the global feature is a rough representation of entire input images, while the instance feature corresponds to object instances, the feature gap results in significant differences in domain probabilities. Existing approaches address this by employing different thresholds and entropy functions, but these methods require manual parameter tuning.  In addition to that, adopting advanced UniDA frameworks such as clustering \cite{saito2020universal,li2021domain}, optimal transport \cite{chang2022unified}, and mutual learning \cite{lu2024mlnet} are complex and challenging to adapt to detection tasks.

\par
To address these issues, we propose a novel Dual Probabilistic Alignment (DPA) framework. For the first issue, we conduct a theoretical analysis to unveil that domain-private alignment is crucial for global-level features. Therefore, we propose a Global-level Domain Private Alignment (GDPA) module that includes global-level sampling, alignment weight calculation, and global-level domain alignment. Global-level sampling aims to mine domain-private category samples. Alignment weight calculation involves the cumulative distribution function to refine the distribution distance estimation as the weight, thereby conducting global-level domain alignment to address the domain-private alignment issue. For the second problem, we conduct a tailored domain-shared category alignment at the instance-level features. To effectively obtain the domain-shared category, we propose a novel unsupervised clustering perspective.  We set the domain label as the center and map samples to domain probabilities to calculate the gradient norm (distance). We then model the frequency of the gradient norm as a Gaussian distribution using bins. The continuous frequency bins of the samples represent those within a certain radius (the sum of bins).  Therefore, we propose an Instance-level Domain Shared Alignment (IDSA) method consisting of instance-level sampling, alignment weight calculation, and instance-level domain alignment. The instance-level sampling utilizes a Gaussian distribution modeling to select domain-shared category samples. Alignment weight calculation involves the Gaussian distribution statistical properties as the weight, thereby conducting instance-level domain alignment to address the heterogeneity of features across different levels. According to the upper bound obtained by theoretical analysis, the PCC module aggregates domain-private category centroids and conducts cross-space consistency in the private category to mitigate negative transfer. In conclusion, our key contributions are as follows:
\begin{itemize}
\item 
We first reveal that domain-specific alignment is crucial for global-level features. Additionally, we provide a theoretical analysis of the upper bound of UniDAOD to support this observation. 
\item A novel unsupervised clustering perspective is proposed to sample the instance samples through continuous frequency bins of the gradient norm as the sampling radius on the Gaussian distribution modeling.
\item We propose a novel Dual Probabilistic Alignment (DPA) framework. DPA aligns domain-private categories at the global and domain-shared categories at the instance level. In addition, the DPA aggregates domain-private categories centroid between feature and probability spaces to mitigate negative transfer.
\item Extensive experiments across open-set, partial-set, and closed-set scenarios demonstrate that the DPA framework achieves state-of-the-art performance, significantly surpassing existing UniDAOD methods.
\end{itemize}

\section{Related Work}
\subsection{Domain Adaptive Object Detection (DAOD)} 
DAOD addresses the covariate shift from labeled data in the source domain to the unlabeled target domain under the closed-set categories. Existing DAOD methods can be categorized into adversarial training and mean teacher paradigms.
As for adversarial training, DAF \cite{chen2018domain} incorporates global and instance alignment modules based on the Faster-RCNN detector, while incremental variants \cite{krishna2023mila} improve global and instance alignment. ATMT \cite{li2023improving} explores the potential of self-supervised learning, and EPM \cite{hsu2020every} introduces a new  FCOS detector. Li et al. propose graph-based alignment methods \cite{li2022scan,li2022sigma} to align class-conditional features, with IGG \cite{li2023igg} enhancing graph generation by effectively addressing non-informative noise. As for the mean teacher paradigm, existing works \cite{chen2022learning,deng2023harmonious,cao2023contrastive,li2023novel,liu2022towards} focus on generating pseudo-labels for the target domain.  In general, existing DAOD works have limited generalizability in open-world scenarios. 

\begin{figure*}[!htbp]
\centering
\includegraphics[width=1.0\textwidth]{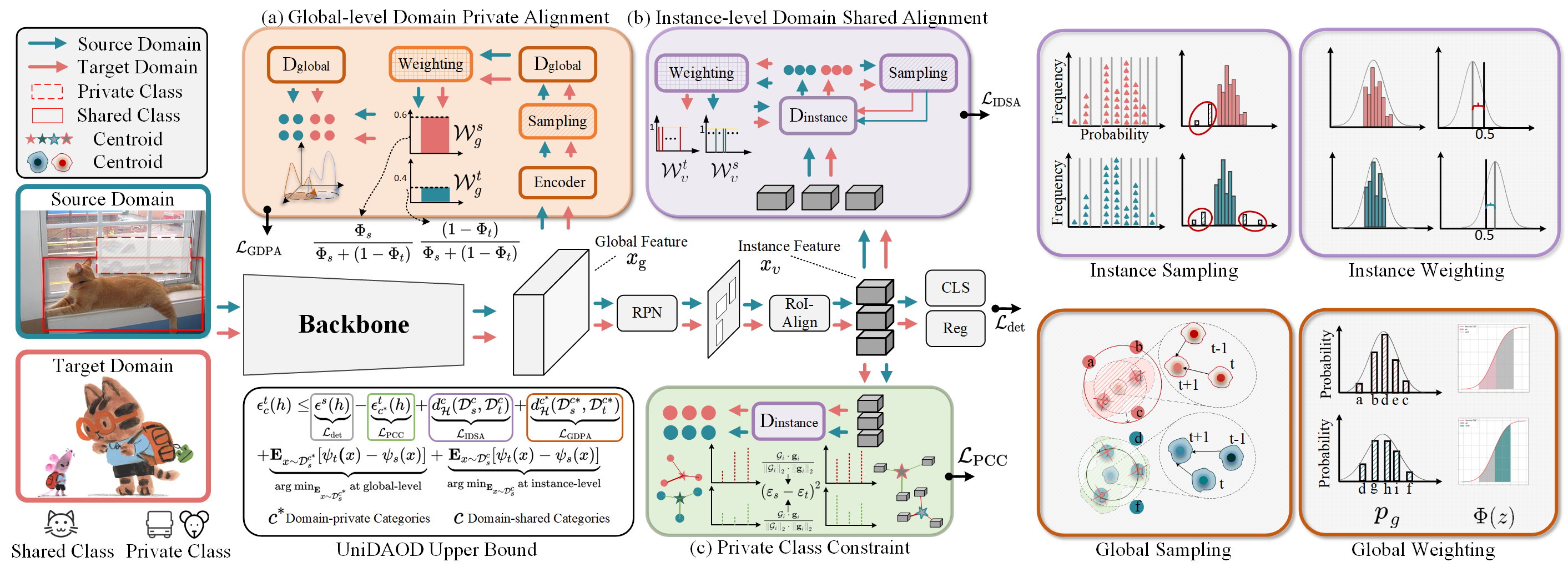}
\caption{
Illustration of the proposed DPA framework. (a) GDPA establishes the global-level embedding feature to sample the outlier in the feature space, then applies a CDF of  Gauss distribution to weighting the probability distribution. (b) IDSA obtains the gradient norm of instance probability and models it as a Gauss distribution for sampling and weighting. (c) PCC obtains the domain-private common centroid and constraints distances of samples to the centroid between feature and probability spaces.
} \label{framework}
\end{figure*}

\subsection{Universal Domain Adaptation (UniDA)}
 UniDA \cite{you2019universal} is a general paradigm for partial-set \cite{zhang2018importance}, open-set \cite{panareda2017open}, and closed-set domain adaptation \cite{tzeng2017adversarial}. The existing UniDA can be categorized into four paradigms, including threshold, clustering, optimal migration, and mutual learning. The threshold methods \cite{you2019universal,fu2020learning,chen2022evidential} estimate inter-sample uncertainty to identify shared categories, often relying heavily on manually set thresholds. Clustering-based UniDA methods \cite{saito2020universal,li2021domain} have been developed to distinguish shared categories. UniOT \cite{chang2022unified} introduces optimal transport to detect shared categories. Despite diversity advancements in classification tasks, UniDAOD remains in its early stages. US-DAF \cite{shi2022universal} conducts scale domain alignment and uses thresholds for sample mining. 
Recently, UCF \cite{lang2023class} and W-adapt \cite{shi2024dynamically} employ weighting mechanisms to mitigate the negative transfer of private categories. CODE \cite{shi2024confused} leverages virtual domain labels to avoid domain-private samples alignment. In general, existing UniDAOD methods perform domain-shared category alignment at both the global and instance levels but overlook the alignment of domain-private categories.

\section{Theoretical Motivation}
We theoretically analyze the error risk upper of UniDAOD for domain-shared and domain-private categories based on the theory of Unsupervised Domain Adaptation (UDA) \cite{ben2010theory}.
\\
\textbf{Definition 1. Universal Domain Adaptation (UniDA).}
We define source and target domains with data ${x}_{s/t}$ with distribution $  \left \{ \mathcal{D}_{s/t} |  \mathcal{P}_{{x}\sim \mathcal{D}_{s}} \ne \mathcal{P}_{{x}\sim \mathcal{D}_{t}}\right \} $, and a label function $\psi:{x} \to  \left \{ {c},{c}^\ast \right \}$, where ${c}$ is the domain-shared categories and  ${c}^\ast$ is the domain-private categories. For simplicity, we omit the $s/t$ notation unless explicitly indicated. The goal of UniDA is to train a model $h$ that can minimize the shared categories error risk of target domain $ \epsilon_{c}(h)=\text{min } \mathbf{E}_{({x},\psi({x})) \sim \mathcal{D}_{t}^{c}}\left [ h({x}) \ne \psi({x})  \right ]$.  
\\
\textbf{Definition 2. Private and Shared Category Error Risks.}
 Given the input $x$ draw from the distribution $\mathcal{D}$ with the label function $\psi$, we can get the error risk as follows:
\begin{equation}
\label{integral}
\small
\epsilon(h) = \mathbb{E}_{{x} \sim \mathcal{D}} \left| h({x}) - \psi ({x}) \right| 
= \int_{{x}} \left| h({x}) - \psi({x})  \right| \, \mathcal{P}_{x, \psi} \, {\rm d} x.  
\end{equation}
We can decompose Eq. \eqref{integral} into the domain-shared categories and domain-private categories and let $\mathcal{F}_{h, \psi}$ denote $\left| h({x}) - \psi({x})  \right|$ as follows:
\begin{equation}
\label{error_min}
\small
\epsilon(h) =  \int_{{x}_{c}} \mathcal{F}_{h, \psi} \, \mathcal{P}_{x, \psi} \, {\rm d}x +\int_{{x}_{{c}^\ast}} \mathcal{F}_{h, \psi} \,  \mathcal{P}_{x, \psi}\, {\rm d}x = \epsilon_{c}(h) + \epsilon_{{c}^\ast}(h),
\end{equation}
where $\epsilon_{c}(h)$ and $\epsilon_{{c}^\ast}(h)$ are the error risk for domain-shared and domain-private categories. Then, we define a symmetric hypothesis space $\mathcal{H}$ based on the error risks upper bound of UDA \cite{ben2010theory} and combined with Eq. \eqref{error_min} to obtain the error risks upper bound for UniDAOD as follows:
\begin{equation}
\small
\label{error_all}
\begin{array}{l}
\epsilon_{c}^{t}(h) \leq \underbrace{\epsilon^{s}(h)}_{\mathcal{L}_{\mathrm{det}}}-\underbrace{\epsilon_{c^{*}}^{t}(h)}_{\mathcal{L}_{\mathrm{PCC}}}+\underbrace{d_{\mathcal{H}}^{c}\left(\mathcal{D}_{s}^{c}, \mathcal{D}_{t}^{c}\right)}_{\mathcal{L}_{\mathrm{IDSA}}}+
\underbrace{d_{\mathcal{H}}^{c^\ast}\left(\mathcal{D}_{s}^{c\ast},\mathcal{D}_{t}^{c\ast}\right)}_{\mathcal{L}_{\mathrm{GDPA}}} \\
+\underbrace{\mathbf{E}_{x \sim \mathcal{D}_{s}^{c^{*}}}\left[\psi_{t}\mathbf({x})-\psi_{s}({x})\right]}_{ \operatorname*{arg\,min}_{\mathbf{E}_{x \sim \mathcal{D}_{s}^{c^{*}}}} \text{at global-level}}+\underbrace{\mathbf{E}_{{x} \sim \mathcal{D}_{s}^{c}}\left[\psi_{t}({x})-\psi_{s}({x})\right]}_{ \operatorname*{arg\,min}_{\mathbf{E}_{x \sim \mathcal{D}_{s}^{c}}} \text{at instance-level}  }.
\end{array}
\end{equation}
\\
\textbf{Remark 1.} Existing UniDAOD methods \cite{shi2022universal,lang2023class,shi2024confused,shi2024dynamically} employ the domain shared category domain alignment $d_{\mathcal{H}}^{c}\left(\mathcal{D}_{s}^{c}, \mathcal{D}_{t}^{c}\right)$ for both the global and instance level features, ignore domain-private category alignment $d_{\mathcal{H}}^{c\ast}\left(\mathcal{D}_{s}^{c\ast},\mathcal{D}_{t}^{c\ast}\right)$ and maximize target domain error risk  $\epsilon_{c^{*}}^{t}(h)$ of the domain-private categories. This oversight leads to an increase in the upper bound of the domain-shared categories, denoted as $\epsilon_{c}^{t}(h)$, in the target domain. In UniDAOD, this issue pertains to the global-level domain-private category alignment and the instance-level domain-shared category alignment. This conclusion is consistent with the observations in Fig. \ref{domain-probs}.

\section{Dual Probabilistic Alignment Framework}
\noindent\textbf{Overview.} The proposed DPA framework is depicted in Fig. \ref{framework}. To minimize the upper bound $\epsilon_{c}^{t}(h)$ of the domain-shared category of the target domain, DPA comprises GDPA, IDSA, and PCC to optimal the terms in Eq. \eqref{error_all}. GDPA minimizes the domain-private category $c^*$ domain distribution discrepancy $d_{\mathcal{H}}^{c^\ast}\left(\mathcal{D}_{s}^{c^\ast},\mathcal{D}_{t}^{c\ast}\right)$ for global-level features, and IDSA minimizes domain-shared category domain distribution discrepancy $d_{\mathcal{H}}^{c}\left(\mathcal{D}_{s}^{c}, \mathcal{D}_{t}^{c}\right)$ for instance-level features. Additionally, PCC maximize the domain-private category risk error $\epsilon_{c^{*}}^{t}(h)$ of target domain.

\subsection{Global-level Domain Private Alignment (GDPA)}
To align global-level domain-private category features, we sample outliers in feature space and model the batch samples as Gauss distribution for the cumulative distribution function to estimate the domain distribution as the weights.
\subsubsection{Global-level Sampling.} The global-level sampling process involves constructing the dynamic feature centroid and updating the learnable radius. The embedding feature $x_{e}$ through the encoder of the domain discriminator computes the dynamic feature centroid and the learnable radius. The dynamic feature centroid, denoted as $\textbf{C} = \textbf{M}(y_d)$, is derived from the memory bank $ \textbf{M} \in \mathbb{R}^{2 \times c'}$. The learnable radius applies the softplus activation function to calculate the boundary $d = \log(1 + e^{\nabla(y_d)})$, where $\nabla \in \mathbb{R}^2 $ represents the learnable boundary parameters and $ y_d $ is the domain label. Subsequently, we can calculate the distance from the sample to the feature centroid and perform sampling with the learnable boundaries as follows:
\begin{equation}
\small
\begin{aligned}
 \Omega^{\text{neg}}_g &= \{ i \mid \|x_i - \textbf{C}\|_2 > d \}, \\  
 \Omega^{\text{pos}}_g &= \{ i \mid \|x_i - \textbf{C}\|_2 \le  d \}, \\ 
\end{aligned}
\end{equation}
where $\Omega^{\text{neg}}_g$ represents the negative sample indexs and $\Omega^{\text{pos}}_g$ represents the positive sample indexs.  Finally, we update the feature centroid and the learnable radius. The memory bank $\textbf{M}(y_d)=\textbf{M}(y_d)\cdot \pi+\overline{x}_e\cdot (1-\pi)$ is adjusted using a momentum update as $ \pi =\frac{ \overline{x}_e\cdot \textbf{M}(y_d)}{\left \|   \overline{x}_e\right \|_2\cdot \left \| \textbf{M}(y_d) \right \|_2  }$, where $\overline{x}_e =\frac{1}{n} \sum_{i}^{n} x_{e,i} $ represents the mean of the current batch of embedding features. Additionally, the learnable radius is updated based on the boundary loss $\mathcal{L}_{\text{bound}}$ as follows:
\begin{equation}
\label{upc}
\small
\mathcal{L}_{\text{bound}} = \frac{1}{n} \sum_{i=1}^{n} \epsilon_i \left( d - \left\| x_i - {\textbf{C}} \right\|_2 \right) + (1 - \epsilon_i) \left( \left\| x_i - {\textbf{C}} \right\|_2 - d \right), 
\end{equation}
where $\epsilon_i = \mathbb{I}(i \in \Omega^{\text{neg}}_g)$ is indicator function.  In contrast to the existing UniDAOD threshold methods \cite{shi2024confused,shi2022universal}, GDPA sampling is data-driven adaptive updating. 
\\
\textbf{Calculating Alignment Weight.} We obtain the domain probability of embedding feature $p_g={\mathcal{F}}_{\mathcal{D}_g}({x}_e)$ through the global level domain discriminator $\mathcal{D}_g$.  To model the Gaussian distribution for the probabilities of the current batch, we estimate expectation $\mu_g= \frac{1}{n}\sum_{i}^{n} p_{g, i}$ and variance $\text{ }\sigma_g^2 = \frac{1}{n}\sum_{i}^{n}(p_{g, i}-\mu_g)^2$. After that, we adopt the cumulative distribution function (CDF), which is calculated as the weight for domain alignment as follows:
\begin{equation}
\small
\label{weight}
\Phi (z)=\frac{1}{2} \left [ 1+ {\rm erf}(\frac{z-\mu_g }{\sigma_g\sqrt{2}  } ) \right ], 
\end{equation}
where ${\rm erf(\cdot)}$ is the Gauss error function, and $z$ is the mean of the probability distribution for adversarial training. As illustrated in Fig. \ref{weight_glo}, we observe that as the shared category ratio $\beta=\protect\frac{\mathcal{C}_{s} \cap \mathcal{C}_{t}}{\mathcal{C}_{s} \cup \mathcal{C}_{t}}$ decreases, the weights $\frac{ \Phi_s/(1-\Phi_t)}{ \Phi_s+(1-\Phi_t)}$ exhibit increased scaling to accommodate a substantial domain gap.
\subsubsection{Global-level Domain Alignment.} To achieve global-level alignment, the gradient reversal layer is employed with focal loss as follows:
\begin{equation}
\small
\label{dag}
\begin{aligned}
\mathcal{L}_{\text{GDPA}} &=  -\frac{1}{n_{\text{neg}}} \sum_{i=1}^{n_{\text{neg}}} \Big[ 
\frac{\Phi_s}{ \Phi_s+(1-\Phi_t)} \left( 1 -p_{s,i}\right)^\gamma \log p_{s,i}  \\
&\quad + \frac{(1-\Phi_t) }{ \Phi_s+(1-\Phi_t)}{p^\gamma _{t,i}}\left( 1 - \log p_{t,i}\right) \Big],
\end{aligned}
\end{equation}
where $\gamma$ is the gamma parameter, $p$ is the probalibity of the domain discriminator, and ${n_{\text{neg}}} \in  \Omega^{\text{neg}}_g$ represents the negative samples numbers. 
\begin{table*}[!htbp]
\centering
\setlength{\tabcolsep}{3pt}
\small
\begin{tabular}{l |  c c c c c c c c c c c c c c c| c}
\toprule
Methods          & boat & bottle & bus  & car  & cat  & chair & cow  & table & dog  & horse & motor & person & plant & sheep & sofa & \textbf{mAP}   \\
\midrule
Source Only     & 31.8 & 41.2   & 31.1 & 34.7 & 5.1  & 33.7  & 23.0 & 20.7  & 8.3  & 43.0  & 52.7  & 49.6   & 40.6  & 17.0  & 13.8 & 29.8 \\
DAF  \cite{chen2018domain}             & 37.2 & 38.0   & 26.9 & 35.9 & 2.3  & 35.2  & 24.0 & 28.5  & 4.2  & 33.8  & 54.7  & 59.4   & 58.4  & 13.4  & 17.9 & 31.3 \\ 
MAF  \cite{he2019multi}  & 24.2 & 42.9   & 35.1 & 32.3 & 11.0 & 41.7  & 22.4 & 32.6  & 6.7  & 40.0  & 59.1  & 52.7   & 41.0  & 24.1  & 17.9 & 32.2 \\ 
HTCN  \cite{chen2020harmonizing}        & 25.9 & 47.8   & 36.0 & 32.8 & 11.3 & 39.4  & 51.7 & 18.7  & 10.5 & 40.9  & 56.3  & 57.9   & 49.4  & 21.3  & 20.4 & 34.7 \\ 
UAN   \cite{you2019universal}      & 26.6 & 37.7   & 48.2 & 31.5 & 8.6  & 32.8  & 23.7 & 31.6  & 2.4  & 36.6  & 56.6  & 42.8   & 44.8  & 14.7  & 16.4 & 30.3 \\ 
CMU   \cite{fu2020learning}   & 14.7 & 41.9   & 52.5 & 34.7 & 9.2  & 36.5  & 38.1 & 21.0  & 7.6  & 37.0  & 48.6  & 55.7   & 44.5  & 17.7  & 21.1 & 32.1 \\ 
SFA \cite{wang2021exploring}  & 25.2 & 30.5& 24.3 & 22.1 & 1.6 & 28.3 & 37.5 & 23.4 & 2.6 & 19.9 & 43.1 & 47.0 & 39.0 & 16.8 & 22.0 & 25.5 \\
US-DAF   \cite{shi2022universal}    & 34.9 & 40.8   & 28.9 & 36.4 & 17.7 & 38.4  & 64.6 & 28.0  & 10.3 & 45.8  & 64.5  & 62.5   & 52.1  & 25.8  & 24.8 & 38.4 \\ 
UCF    \cite{lang2023class}     & 36.2 & 44.3   & 28.3 & 37.1 & 2.2  & 36.0  & 61.9 & 27.7  & 4.0  & 39.9  & 64.7  & 64.2   & 52.6  & 20.9  & 26.9 & 36.5 \\ 
CODE  \cite{shi2024confused} & 36.8 & 45.1& 42.0&37.7& 18.4& 44.5& 47.7& 33.1& 8.4& 45.7& 69.2& 61.7& 50.7& 25.5& 24.6& 39.4 \\
\midrule
DPA (Ours)  &      32.9 & 46.0 & 62.1 & 41.4 & 4.2 & 42.0 & 64.4 & 33.3 & 8.1 & 40.5 & 67.1 & 64.2 & 57.8 & 32.1 & 25.1 & \textbf{41.4}    \\ 
  \bottomrule
  \end{tabular}
\caption{Comparison on Pascal VOC to Clipart1k (Open-set: $\beta= 75\%, C_s \cap C_t \neq \emptyset,  C_s  \setminus C_t\neq \emptyset,C_t \setminus C_s \neq \emptyset$ ).}
\label{voc075}
\end{table*}

\begin{table*}[!ht]
\centering
\begin{tabular}{l| c c c c c c c c c c |c}
\toprule
Methods          & bus  & car  & cat  & chair & cow  & table & dog & horse & motor & person & \textbf{mAP}  \\
\midrule
Source Only      & 43.3 & 33.0 & 8.4  & 32.1  & 24.0 & 28.7  & 6.9 & 34.9  & 51.8  & 42.5   & 30.6 \\
DAF  \cite{chen2018domain}             & 37.5 & 32.8 & 10.2 & 40.3  & 27.2 & 31.3  & 4.1 & 41.0  & 55.5  & 52.0   & 33.2 \\
MAF  \cite{he2019multi}              & 37.1 & 31.1 & 9.7  & 38.1  & 19.9 & 29.1  & 2.5 & 37.3  & 50.7  & 50.0   & 30.6 \\
HTCN  \cite{chen2020harmonizing}              & 29.5 & 34.4 & 17.3 & 33.8  & 50.6 & 14.0  & 3.6 & 46.9  & 74.7  & 58.5   & 36.3 \\
UAN   \cite{you2019universal}                  & 48.9 & 26.4 & 14.6 & 36.7  & 49.9 & 30.0  & 3.2 & 39.9  & 56.1  & 52.0   & 35.8 \\
CMU   \cite{fu2020learning}               & 33.3 & 32.8 & 8.1  & 41.5  & 55.5 & 24.6  & 5.6 & 43.3  & 54.9  & 60.4   & 36.0 \\
SFA \cite{wang2021exploring}              & 29.4 & 28.3 & 14.4 & 30.5  & 29.4 & 13.0  & 3.1 & 26.8  & 61.6  & 2.9    & 27.9\\
US-DAF   \cite{shi2022universal}           & 31.3 & 41.9 & 7.3  & 42.9  & 64.3 & 30.0  & 5.7 & 44.8  & 69.5  & 61.9   & 40.0 \\
UCF    \cite{lang2023class}             & 32.4 & 37.4 & 4.0  & 33.5  & 59.3 & 40.2  & 4.4 & 39.0  & 58.1  & 61.7   & 37.0 \\
CODE  \cite{shi2024confused}              & 50.0 & 38.8 & 19.5 & 42.0  & 46.9 & 34.5  & 13.5& 43.7  & 64.3  & 52.9   & 40.6 \\
\midrule
DPA (Ours)  &      45.0 & 41.3 & 13.7 & 38.0 & 64.9 & 30.0 & 13.8 & 45.8 & 74.0 & 60.2 &  \textbf{42.7}    \\
\bottomrule
\end{tabular}
\caption{Comparison on Pascal VOC to Clipart1k (Open-set: $\beta=50\%, C_s \cap C_t \neq \emptyset,  C_s  \setminus C_t \neq \emptyset,C_t \setminus C_s \neq \emptyset$). }
\label{voc05}
\end{table*}
\subsection{Instance-level Domain Shared Alignment (IDSA)} 
To efficiently align the domain-shared category at the instance level, we calculate the gradient norm of the instance samples to model a Gaussian distribution to discard outlier samples and estimate the weight of the domain-shared category samples to improve domain alignment.
\\
\textbf{Instance-level Sampling.} The instance-level sampling process involves constructing the probability space and the sampling criteria. First, we build the probability space to model the Gaussian distribution. The number of $\hat{n}$ instance-level features $x_\upsilon$ is generated by the domain discriminator ${\mathcal{F}}_{\mathcal{D}_\upsilon}$ to calculate the domain probabilities $p_\upsilon  = {\rm Sigmoid} \left( {\mathcal{F}}_{\mathcal{D}_\upsilon}(x_\upsilon) \right)$. The domain probabilities are used to compute the gradient norm $\eta_\upsilon $ for each instance as $\eta_\upsilon = \left|  p_\upsilon- {y_d} \right|$, where $ {y_d}$ is the domain label. We then construct the gradient norm bins $\hat{\Omega}=\left\{ i\cdot\psi | i \in \mathbb{Z} \right\}$  to calculate the gradient norm frequencies $\tau$, which model the Gaussian distribution with the minimum interval $\psi=
\mathop {\rm argmin }\left\{ (\eta_\upsilon^{\rm max}-\eta_\upsilon^{\rm min})\cdot\eta_\upsilon^{\rm std},\delta  \right\} $, where $\delta$ is the hyperparameter. For the sampling criteria, we leverage the statistical characteristics of the Gaussian distribution. The first sampling criterion involves filtering out samples in noncontinuous frequency bins. The second criterion is related to the characteristic of the Gaussian distribution, where the frequency of filtered bins is lower than that of continuous bins. The sampling process is as follows:
\begin{equation}
\small
\begin{aligned}
&{\rm \Omega}^{\rm pos}_\upsilon = \left\{ i \mid \tau_i > 0, \; i_{j+1} - i_j = 1  ,\forall j  \right\}, \\
&{\rm \Omega}^{\rm neg}_\upsilon = \left\{ i \mid i \notin {\rm \Omega}^{\rm pos}_\upsilon, \; \tau_i < \tau_\omega\right\},
\end{aligned}
\end{equation}
where $\tau_\omega$ denotes the frequency in the first and last continuous bins in positive samples ${\rm \Omega}^{\rm pos}_\upsilon $. The sum of the bins represents the sampling radius from the feature centroid in the feature space, which dynamically adjusts the bins following a Gaussian distribution of the source or target domain data during adversarial training.
\\
\textbf{Calculating Alignment Weight.} These negative instances are excluded from the instance alignment through the instance weight as follows:
\begin{equation}
\small
\mathcal{W}_\upsilon = \left\{
\begin{array}{ll}
0, & \mathcal{W}_\upsilon \in {\rm \Omega}^{\rm neg}_\upsilon, \\ 1-\frac{\left | \eta_\upsilon^{\rm mean}-0.5 \right |}{0.5}  , & \mathcal{W}_\upsilon \in {\rm \Omega}^{\rm pos}_\upsilon,
\end{array}
\right.
\end{equation}
where $\eta_\upsilon^{\rm mean}$ is the mean value of the gradient norm $\eta_\upsilon$.
\subsubsection{Instance-level Domain Alignment.} This processing aims to provide instance-level features into the domain discriminator for adversarial training to achieve domain alignment. Based on the obtained weights $\mathcal{W}_\upsilon$, the loss function of the IDSA module is as follows:
\begin{equation}
\small
\mathcal{L}_{\rm IDSA} =  -\frac{1}{\hat{n}}\sum\limits_i^{\hat{n}} \mathcal{W}_\upsilon \cdot(1-p_\upsilon){\rm log}(p_\upsilon) + \mathcal{W}_\upsilon \cdot p_\upsilon(1-{\rm log}(p_\upsilon)),
\end{equation}
where $\hat{n}$ is the number of instance proposals. By optimizing the function ${\mathcal{L}_{\rm IDSA}}$ in adversarial training, the IDSA module mitigates negative transfer caused by domain-private feature alignment. It calibrates the domain-shared feature distribution according to a Gaussian distribution to enhance positive transfer between source and target domains.

\subsection{Private Class Constraint (PCC)} 
Given the instance-level feature $x_\upsilon$ and the domain probabilities $p_\upsilon$, we first perform a classifier head to establish domain-private categories for both the source domain and the target domain: $ \{ {c}^\ast_i \mid \hat{y}_{s, i}^{{c}^\ast} \cap \hat{y}_{t, i}^{{c}^\ast} = \emptyset \}^{n^*}_i $. To aggregate the centroids of domain-private categories in feature and probability spaces, we calculate the feature centroid $ \bar{x}_\upsilon $ and the probability centroid $ \bar{p}_\upsilon$. We then conduct the cosine similarity distance $ \mathcal{G}_i = \left\| x_{\upsilon, i} - \bar{x}_\upsilon \right\|_2 $ and probability samples $ \mathbf{g}_i = \left\| p_{\upsilon, i} - \bar{p}_\upsilon \right\|_2 $ to centroid to measure intra-domain distances. To measure the intra-domain distance, we use cosine similarity, defined as $ \varepsilon_{s/t} = \frac{1}{n^*} \sum_{i=1}^{n^*} \frac{\mathcal{G}_i \cdot \mathbf{g}_i}{\left\| \mathcal{G}_i \right\|_2 \cdot \left\| \mathbf{g}_i \right\|_2} $. Finally, we employ the mean squared error (MSE) loss function to minimize the inter-domain distance, as follows:
\begin{equation}
\small
{\mathcal L_{\rm PCC}}={ {{{\left( { \varepsilon_s - \varepsilon_t} \right)}^2}} }.
\end{equation}
The loss function ${\mathcal L_{\rm PCC}}$ optimizes the network by adopting a gradient detach for the source domain.

\subsection{Optimization}
The training loss of the DPA is represented as ${\mathcal L}_{\rm DPA}$, which consists of the following loss terms:
\begin{equation}
{\mathcal L}_{\rm DPA} =  {\mathcal L_{\rm det}} + {\mathcal L_{\rm GDPA}} +{\mathcal L_{\rm IDSA}} +\alpha {\mathcal L_{\rm PCC}},
\end{equation} 
where $\mathcal{L}_{\rm det}$ is the Faster-RCNN detector loss. $\mathcal{L}_{\rm GDPA}$ and $\mathcal{L}_{\rm IDSA}$ are the domain alignment losses for the GDPA and IDSA modules at the global and instance levels, respectively.  The ${\mathcal L}_{\rm DPA}$ is optimized using the SGD optimizer. The bound loss ${\mathcal{L}_{\rm{bound}}}$ is optimized using the Adam optimizer with a learning rate set to $0.1$. The hyperparameter of $\alpha$ is 0 for the initial epoch and 0.1 thereafter.

\section{Experiments} 
\begin{table}[!ht]
\centering
 \setlength{\tabcolsep}{1mm}
\begin{tabular}{l| c c c c c |c}
\toprule
 Methods & plane & bicycle & bird & boat & bottle & \textbf{mAP} \\
\midrule
Source Only  & 33.2  & 55.7    & 25.4 & 29.2 & 41.6   & 37.0 \\
DAF      & 31.5  & 42.5    & 25.2 & 34.4 & 50.8   & 36.9 \\
MAF              & 29.3  & 57.0    & 27.1 & 33.9 & 41.8   & 37.8 \\
HTCN             & 32.5  & 53.0    & 24.1 & 27.0 & 48.4   & 37.0 \\
UAN        
              & 35.6  & 55.9    & 27.1 & 28.2 & 44.2   & 38.2 \\
CMU                & 45.5  & 52.7    & 28.8 & 29.4 & 40.1   & 39.3 \\
SFA     & 28.4  & 32.4    & 27.2 & 34.2 & 34.2   & 31.3 \\
US-DAF                & 44.2  & 57.5    & 27.9 & 32.2 & 40.5   & 40.5 \\
UCF                & 35.8  & 52.9    & 28.6 & 20.8 & 55.7   & 38.8 \\
CODE                & 42.1  & 61.4    & 26.2 & 32.1 & 44.1   & 41.2 \\
\midrule
DPA (Ours)           & 40.8 & 58.1 & 28.2 & 33.7 & 52.0 & \textbf{42.5}    \\
\bottomrule
\end{tabular}
\caption{Comparison on Pascal VOC to Clipart1k (Open-set: $\beta=25\%, C_s \cap C_t \neq \emptyset,  C_s  \setminus C_t\neq \emptyset,C_t \setminus C_s \neq \emptyset$). }
\label{voc25}
\end{table}

\begin{table}[!ht]
\centering
\setlength{\tabcolsep}{2pt}
\begin{tabular}{l| c c c c c c |c}
\toprule
Methods     & bicycle & bird & car  & cat  & dog  & person & \textbf{mAP}  \\
\midrule
Source Only      & 29.8    & 50.2 & 47.1 & 62.2 & 51.5 & 57.8   & 49.8 \\
DAF           & 29.5    & 53.8 & 50.6 & 58.1 & 48.1 & 56.5   & 49.4 \\
MAF             & 28.5    & 50.0 & 46.8 & 59.4 & 50.2 & 58.6   & 48.9 \\
HTCN            & 26.4    & 43.0 & 46.5 & 50.8 & 44.0 & 53.9   & 44.1 \\
UAN             & 33.6    & 52.1 & 53.8 & 62.4 & 52.2 & 56.1   & 51.7 \\
CMU        & 36.9    & 51.2 & 53.3 & 59.3 & 51.7 & 59.9   & 52.0 \\
CODE             & 39.6    & 53.1 & 54.7 & 57.6 & 56.1 & 57.8   & 53.1 \\
SFA        &  4.4 &  17.4  & 11.1 & 14.9 & 20.1 & 18.4 & 14.4 \\
US-DAF          & 35.0    & 52.4 & 52.7 & 63.1 & 54.3 & 59.8   & 52.9 \\
UCF                  & 34.8    & 52.0 & 53.8 & 61.9 & 54.2 & 60.5   & 52.9 \\
\midrule
DPA (Ours)  & 31.3 & 55.2 & 56.4 & 61.1 & 62.1 & 58.3 & \textbf{54.1} \\
\bottomrule
\end{tabular}
\caption{
Comparison WaterColor to Pascal VOC (Partial-set: $\beta=30\%, C_s \subset  C_t$).}
\label{w2p}
\end{table}
\begin{table}[!ht]
\centering
\setlength{\tabcolsep}{2pt}
\begin{tabular}{l| c c c c c c |c}
\toprule
Methods          & bicycle & bird & car  & cat  & dog  & person & \textbf{mAP}  \\
\midrule
Source Only      & 82.4    & 51.7 & 48.4 & 39.9 & 30.7 & 59.2   & 52.0 \\
DAF            & 73.4    & 51.9 & 43.1 & 35.6 & 28.8 & 63.1   & 49.3 \\
MAF              & 70.4    & 50.3 & 44.3 & 36.7 & 30.6 & 62.9   & 49.2 \\
HTCN            & 74.1    & 49.8 & 51.9 & 35.3 & 35.3 & 66.0   & 52.1 \\
UAN                & 78.0    & 53.6 & 50.4 & 36.4 & 35.8 & 65.6   & 53.3 \\
CMU               & 82.0    & 53.9 & 48.6 & 39.6 & 33.1 & 66.0   & 53.9 \\
SFA        & 37.1    & 39.3 & 32.3 & 52.7 & 9.9  & 34.1   & 34.2 \\
US-DAF            & 86.5    & 54.1 & 50.0 & 43.0 & 34.0 & 63.2   & 55.2 \\
UCF           & 84.8    & 52.1 & 49.8 & 40.6 & 33.8 & 63.2   & 54.1 \\
CODE         & 87.9    & 55.3 & 50.7 & 38.9 & 34.7 & 67.5   & 55.8 \\
\midrule
DPA (Ours)         &  86.1 & 53.5 & 50.3 & 44.1 & 38.1 & 65.6 &  \textbf{56.3}  \\
\bottomrule
\end{tabular}
\caption{Comparison on Pascal VOC to WaterColor (Partial-set: $\beta=30\%$, ${C_s} \supset {C_t}$). }
\label{p2w}
\end{table}
\begin{table}[!ht]
\centering
\setlength{\tabcolsep}{2pt}
\begin{tabular}{ l c c c}
\toprule
Methods   & DA Settings  & \textbf{mAP}	  \\
\midrule
SFA \cite{wang2021exploring} &DAOD  &41.3 \\
SCAN++ \cite{li2022scan++}  &DAOD &42.8   \\
SIGMA++ \cite{li2023sigma++} &DAOD &44.5 \\
 US-DAF \cite{shi2022universal} &UniDAOD & 37.8  \\
 UCF \cite{lang2023class}  &UniDAOD  & 34.2  \\
% W-adapt \cite{shi2024dynamically} &UniDAOD & 42.5 \\
CODE \cite{shi2024confused}  &UniDAOD & 42.1  \\
\midrule
DPA (Ours) &UniDAOD   & \textbf{46.3}   \\
\bottomrule
\end{tabular}
\caption{ Comparison on Cityscape to Foggy Cityscape (Closed-set: $\beta=100\%, {C_s} = {C_t}$).}
\label{city}
\end{table}
\subsection{Implementation Details} 
We conduct extensive experiments following the setting \cite{shi2022universal} for three benchmarks: open-set, partial-set, and closed-set. The baseline approach \cite{chen2018domain} adopts Faster-RCNN as the base detector with the focal loss, and the backbone is ResNet-101 \cite{he2016deep} or VGG-16 in \cite{simonyan2014very} pre-trained on ImageNet \cite{deng2009imagenet}, which adopt VGG-16 in Cityscape to Foggy Cityscape and other benchmarks adopt the ResNet-101. The DPA model optimized training iterations are $100$k, with an initial learning rate of 1e-3 and a subsequent decay of the learning rate to 1e-4 following $50$k iterations. The detection performance is evaluated with the mean Average Precision (mAP) metric, and the threshold of mAP follows the setting \cite{shi2022universal} to set 0.5.

\subsection{Datasets and Domain Adaptation Settings} 
We evaluate our DPA framework on five datasets across three domain adaptation scenarios (open-set, partial-set, and closed-set): Foggy Cityscapes \cite{sakaridis2018semantic}, Cityscapes \cite{cordts2016cityscapes}, Pascal VOC \cite{everingham2010pascal}, Clipart1k \cite{inoue2018cross}, and Watercolor \cite{inoue2018cross}. In the open-set scenario, there are shared and private categories in both the source and target domains. We introduce mutil ratios $\beta=\left\{0.25, 0.5, 0.75\right\}$ to construct different shared category ratios  $\beta=\protect\frac{\mathcal{C}_{s} \cap \mathcal{C}_{t}}{\mathcal{C}_{s} \cup \mathcal{C}_{t}}$ benchmarks. In the partial-set scenario,  the category set of the source domain is the subset for the target domain, and vice versa. In the closed-set scenario, the categories in the target and source domains are identical.

\subsection{Comparisons with the State-of-the-Arts}
\textbf{Open-set scenario.} Tables \ref{voc075}, \ref{voc05}, and \ref{voc25} present the open-set domain adaptive object detection performance from Pascal VOC to Clipart1k under different category overlap ratios $\beta$. The proposed framework consistently achieves state-of-the-art performance across various category settings compared to other methods. Compared to related DAOD methods (HTCN, MAF, DAF, SFA), the proposed DPA framework demonstrates significant performance advantages. Additionally, in comparison with UniDAOD methods (CODE, US-DAF, CMU, UAN, and UCF), the proposed DPA framework also exhibits superior performance. \\
\textbf{Partial-set scenario.} The results of the partial-set domain adaptation are presented in \ref{w2p} and \ref{p2w}. In the partial-set scenario, the private categories are exclusively present in the source or target domain, leading to negative transfer. The proposed DPA method effectively addresses this issue and outperforms other approaches by a significant margin, achieving 54.1\% and 56.3\% mAP. For the domain-private category in the target domain ($C_s \subset  C_t$),  DPA enhances performance by 1.2\% compared to the UniDAOD method (US-DAF), as shown in Table \ref{w2p}. For the domain-private category in the source domain (${C_s} \supset {C_t}$), DPA improves performance by 4.2\% compared to the DAOD method (HTCN).
% , as shown in Table \ref{p2w}.
\\
\textbf{Closed-set scenario.} The closed-set scenario results are shown in Table \ref{city} show that although existing UniDAOD methods \cite{shi2024confused,shi2024dynamically}  integrated with advanced DAOD methods achieve notable performance improvements on closed-set, the DAOD methods significantly outperform the UniDAOD methods. This advantage comes from DAOD under the closed-set assumption, while UniDAOD prioritizes dealing with label shifts in open environments. The proposed DPA framework exhibits satisfactory performance in closed-set scenarios through probability modeling.

\subsection{Ablation Study} 
We conduct ablation experiments on each submodule, with the corresponding results presented in Table \ref{abl}.  In these experiments, each module of the proposed DPA framework improves performance. The GDPA and IDSA modules provide significant gains when the domain-shared categories ratio is high ($\beta=50\%, 75\% $), while the PCC module leads to more substantial improvements when the ratio is low ($\beta=25\%$).
\begin{table}[!ht]
\centering
\begin{tabular}{ l c c c }
\toprule
Ratio $\beta$ & $75\%$ & $50\%$ & $25\%$   \\
\midrule
Baseline        & 33.9  & 37.9	& 38.2    \\
DPA \textit{w/o} GDPA & 38.3  & 39.7 &  40.1   \\
DPA \textit{w/o} IDSA & 36.3  & 40.1 & 39.9   \\
DPA \textit{w/o} PCC  & 40.7 & 42.4  & 39.4   \\
DPA (Ours)            & \textbf{41.4}   & \textbf{42.7} &  \textbf{42.5}  \\
\bottomrule
\end{tabular}
\caption{Ablation study on Pascal VOC to Clipart1k (Open-set: $\beta=75\%, 50\%, 25\%$).
}
\label{abl}
\end{table}
\subsection{Category-wise Performance Analysis}
\begin{figure}[!ht]
    \centering
\includegraphics[width=0.82\linewidth]{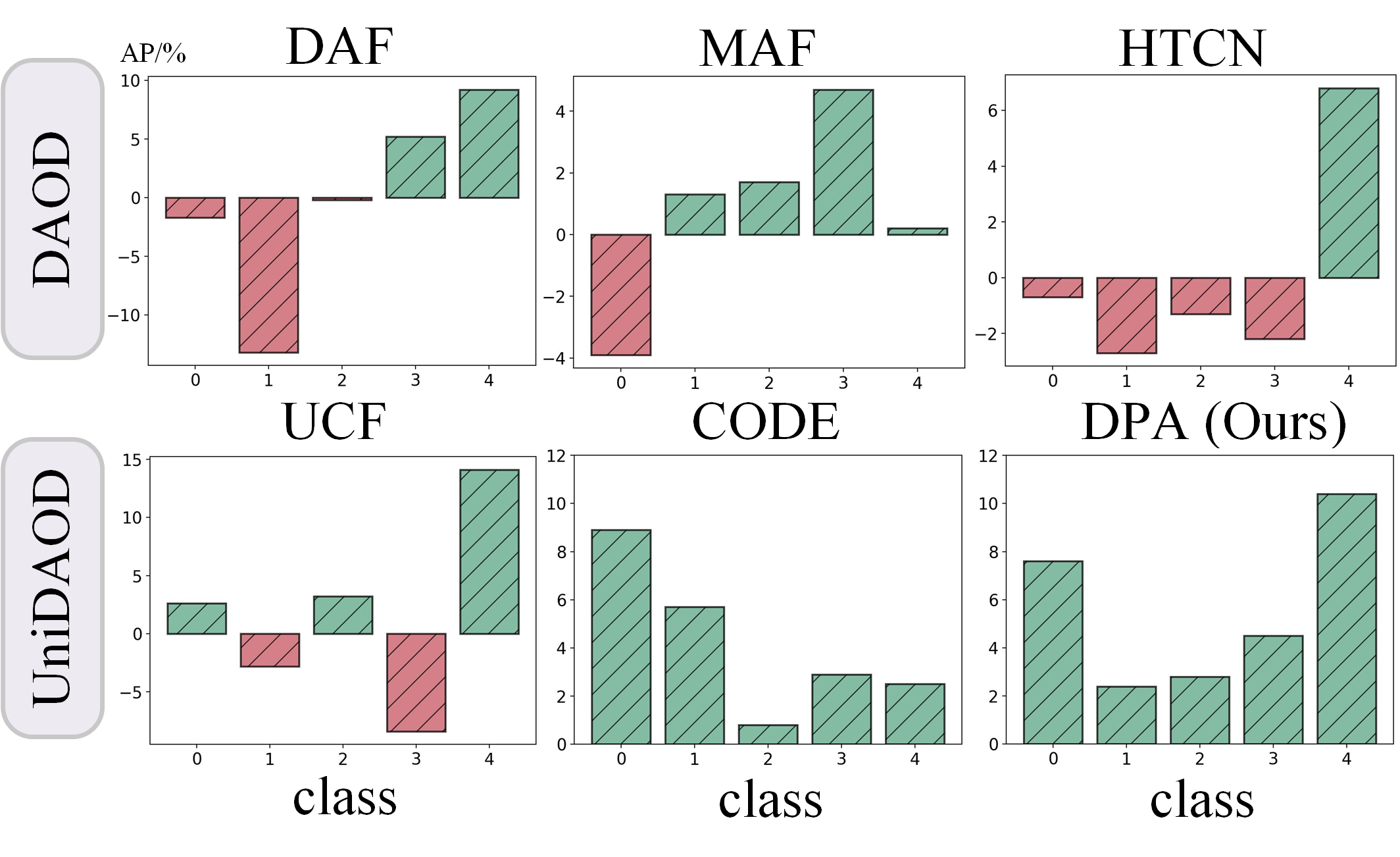}
    \caption{Category-wise performance gain over the source-only model (classes are plane, bicycle, bird, boat, and bottle). \textcolor[HTML]{5ba585}{Positive transfer} is green, and 
   \textcolor[HTML]{ce6772}{negative transfer} is red. }
    \label{acc_gain}
\end{figure}

To compare the performance of the proposed method with existing DAOD and UniDAOD methods in terms of positive and negative transfer, we present the performance gains of DAOD and UniDAOD relative to the source-only model in Fig. \ref{acc_gain}. The DAOD methods exhibit significant negative transfer, where DAF, MAF, and HTCN drop by approximately 2\%, 4\%, and 1\% AP in class 0, respectively. In contrast, the UniDAOD methods mitigate negative transfer, with CODE and DPA achieving positive transfer of around 3\% and 10\% in class 4, respectively. This category-wise performance analysis proves that the proposed method effectively combats negative transfer and strengthens positive transfer.

\begin{figure}[!ht]
    \centering
\includegraphics[width=1.0\linewidth]{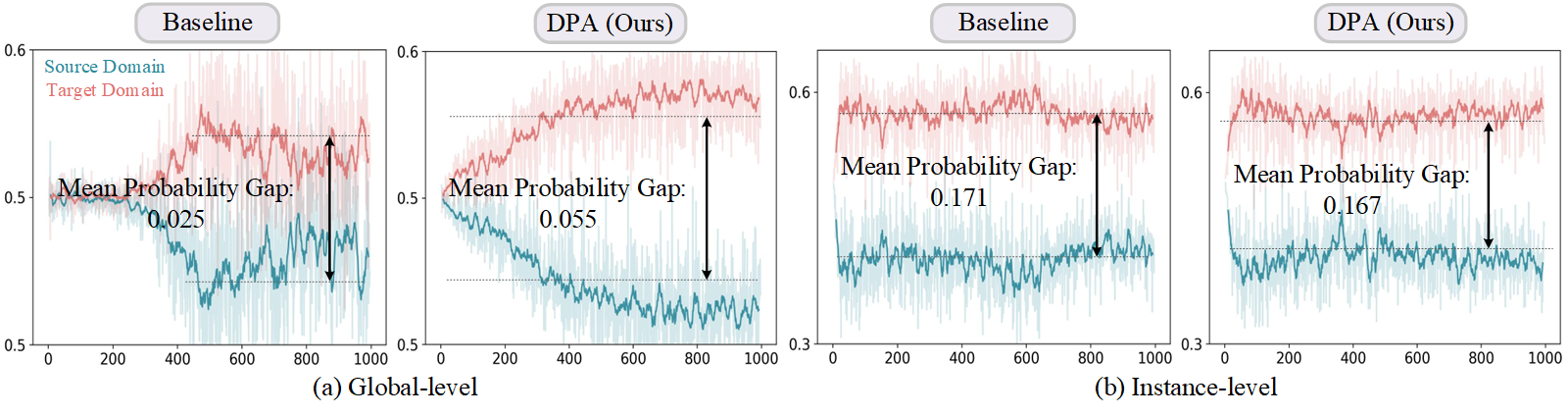}
    \caption{ Qualitative analysis of category alignment in terms of the mean probability gap: (a) global-level features and (b) instance-level features. The horizontal axis represents training iterations ($\times$100), and the vertical axis shows the probability of the domain discriminator. The benchmark is Pascal VOC to Clipart1k. ($\beta=25\%$). }
    \label{last_prob}
\end{figure}
\begin{figure}[!ht]
    \centering
\includegraphics[width=1.0\linewidth]{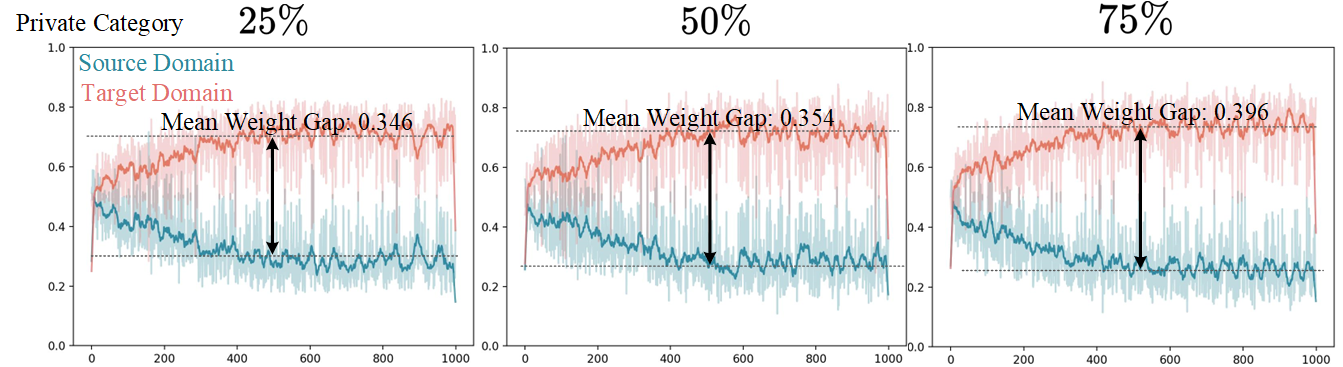}
    \caption{The weight quantitative analysis of global-level domain-private feature. The horizontal axis is training iteration ($\times$100), and the vertical axis is weight values $\frac{ \Phi_s/(1-\Phi_t)}{ \Phi_s+(1-\Phi_t)}$ in source and target domains.}
    \label{weight_glo}
\end{figure}

\subsection{Qualitative Open-set Alignment Analysis}
We further analyze the probability gap in our DPA framework for open-set alignment. As shown in Fig. \ref{last_prob}(a), the global-level mean probability gap is more pronounced in our DPA, highlighting its effectiveness in distinguishing domain-private categories. In contrast, Fig. \ref{last_prob}(b) shows a smaller mean probability gap at the instance level, demonstrating that our DPA better aligns domain-shared categories. Additionally, we perform a weight quantitative analysis of global-level domain-private alignment, as illustrated in Fig. \ref{weight_glo}. As the ratio of domain-private categories increases, the mean weight gap also increases, indicating that adversarial training adaptively penalizes features associated with domain-private categories through weight adjustments. 

\section{Conclusion}
We propose a DPA framework for universal domain adaptive object detection with two kinds of probabilistic alignment. Inspired by a theoretical perspective, we propose a GDPA module for aligning global-level private samples and an IDSA module for aligning instance-level domain-shared samples. To combat negative transfer,  we propose a PCC module to confuse the discriminability of private categories. Extensive experiments are conducted on open, partial, and closed set scenarios and demonstrate our DPA outperforms state-of-the-art UniDAOD methods by a remarkable margin.
\section{Acknowledgments}
This work is supported by the InnoHK program, and the National Natural Science Foundation of China (Grant No.\#62306313).

\bibliography{aaai25}

\end{document}